\documentclass[conference]{IEEEtran}
\IEEEoverridecommandlockouts

\usepackage[hidelinks]{hyperref}
\usepackage{cite}
\usepackage{amsmath,amssymb,amsfonts}
\usepackage{graphicx}
\usepackage{textcomp}
\usepackage{xcolor}

\usepackage{algorithm}
\usepackage{color}
\usepackage{algpseudocode}
\usepackage{mathtools}
\usepackage{amsmath}
\usepackage{amssymb}
\usepackage{subcaption}
\usepackage[flushleft]{threeparttable}
\usepackage{booktabs}
\usepackage{multirow}
\usepackage{marvosym}

\def\BibTeX{{\rm B\kern-.05em{\sc i\kern-.025em b}\kern-.08em
    T\kern-.1667em\lower.7ex\hbox{E}\kern-.125emX}}

\begin{document}

\title{
Robust and Noise-resilient Long-Term Prediction of Spatiotemporal Data Using Variational Mode Graph Neural Networks with 3D Attention}

\author{	\IEEEauthorblockN{Osama Ahmad, Zubair Khalid}
	\IEEEauthorblockA{\textit{School of Science and Engineering, Lahore University of Management Sciences},
		 Lahore 54792, Pakistan\\
	osama\_ahmad@lums.edu.pk, \,zubair.khalid@lums.edu.pk}
}

\maketitle

\begin{abstract}
This paper focuses on improving the robustness of spatiotemporal long-term prediction using a variational mode graph convolutional network (VMGCN) by introducing 3D channel attention. The deep learning network for this task relies on historical data inputs, yet real-time data can be corrupted by sensor noise, altering its distribution. We model this noise as independent and identically distributed (i.i.d.) Gaussian noise and incorporate it into the LargeST traffic volume dataset, resulting in data with both inherent and additive noise components. Our approach involves decomposing the corrupted signal into modes using variational mode decomposition, followed by feeding the data into a learning pipeline for prediction. We integrate a 3D attention mechanism encompassing spatial, temporal, and channel attention. The spatial and temporal attention modules learn their respective correlations, while the channel attention mechanism is used to suppress noise and highlight the significant modes in the spatiotemporal signals. Additionally, a learnable soft thresholding method is implemented to exclude unimportant modes from the feature vector, and a feature reduction method based on the signal-to-noise ratio (SNR) is applied. We compare the performance of our approach against baseline models, demonstrating that our method achieves superior long-term prediction accuracy, robustness to noise, and improved performance with mode truncation compared to the baseline models. The code of the paper is available at 
\underline{\texttt{https://github.com/OsamaAhmad369/VMGCN}}.
\end{abstract}

%

\section{Introduction}
Time series analysis involving spatial data has rapidly expanded across various fields, including arrhythmia detection using electrocardiogram (ECG)~\cite{zhang2020ecg}, traffic accident prediction~\cite{zhu2019ta},  traffic flow prediction~\cite{ali2021exploiting}, flood forecasting~\cite{ding2020interpretable}, and air quality prediction~\cite{huang2021spatio}. In forecasting applications, historical data is fed into deep learning algorithms to learn complex patterns. This historical data is gathered from the sensors, which can sometimes be corrupted, potentially affecting the prediction accuracy of the neural networks. This work examines the robustness of models against noise in real-time traffic data and aims to improve long-term prediction accuracy. 

Recent research has proposed state-of-the-art modeling techniques for better understanding and modeling complex spatial patterns and temporal dynamics. Graph neural networks (GNNs) are frequently used in learning models to capture the spatial correlation within the graph networks~\cite{li2017diffusion},~\cite{cai2020traffic}. The gated recurrent units (GRUs)~\cite{chung2014empirical} and transformers~\cite{vaswani2017attention} are powerful tools for dynamically mapping the time-sequential events. An adaptive graph is generated from dynamic node attributes and integrated with a static graph to explore the topology of the graph network~\cite{li2023dynamic}. External factors such as time, point of interest (POI), and weather information, are used to extract valuable insights.  Signal processing-based noise removal methods are also employed in spatiotemporal forecasting problems.

One study applied complete ensemble empirical mode decomposition with adaptive noise (CEEMDAN) to time series data, followed by a stacked autoencoder for feature extraction and long short-term memory (LSTM) for temporal modeling~\cite{liu2019ensemble}. To reduce the effect of the noise in time series, Li et al. implemented the bi-directional long short-term memory (BILSTM) in combination with distinct decomposition methods such as wavelet (WL), empirical mode decomposition (EMD), and ensemble empirical mode decomposition (EEMD)~\cite{li2022traffic}. Variational mode decomposition (VMD) and LSTM have also been employed for short-term traffic flow prediction, though these methods primarily focus on temporal data, neglecting spatial relationships~\cite{lu2023efficient}. However, since these methods primarily focus on temporal aspects, they neglect spatial relationships essential to accurate forecasting.

In air quality prediction, an EEMD-CEEMDAN-GCN (graph convolutional network) framework was proposed to capture spatial features, though it struggles with learning sequential dynamic trends and introduces high computational complexity due to iterative algorithms~\cite{bhatti2024aiot}. Sun et al. proposed ModWaveMLP, where noise from traffic data is removed using a wavelet decomposition module~\cite{sun2024modwavemlp}. Fuzzy clustering has also been applied to residual time series, combined with a convolutional-LSTM for long-term spatial pattern identification, though this model does not account for traffic network topology~\cite{asadi2020spatio}. Another method, referred to as the variational mode graph convolution network (VMGCN), has been developed to decompose temporal graph data into distinct modes for future state prediction using a neural network~\cite{Ahmad2024variational}. Although the model claims to differentiate between modes based on noise levels, substantial evidence for effective noise handling is lacking, and its performance diminishes for long-term predictions compared to short-term forecasts.

Existing decomposition methods, such as EMD, EEMD, and CEEMDAN-based approaches suffer from mode-mixing, while VMD overcomes these challenges by ensuring better frequency separation and adaptability and therefore, offers improved forecasting performance~\cite{zhao2023hybrid}. In this work, we make the following novel contributions to address the aforementioned challenges:
\begin{itemize}
    \item We propose a framework that decomposes the noise-corrupted data into modes and applies 3-D attention, graph convolution, and time convolution to estimate future time events. The 3-D attention mechanism consists of three types of attention: spatial, temporal, and channel. To the best of our knowledge, this is the first time channel attention has been applied to distinguish between modes containing valuable information and those containing noise. 
    \item We conduct experiments using LargeST, a benchmark dataset for traffic flow prediction, which encompasses a wide range of spatial and time data to demonstrate the better prediction performance of the proposed architecture. We use additive noise in spatio-temporal data to train the model and analyze the impact of noise on the prediction accuracy. We also analyze various case scenarios to assess the impact of different hyper-parameters on the performance of our network. Through different experiments, we demonstrate our approach delivers better long-term prediction accuracy, enhanced noise robustness, and superior performance with mode truncation compared to the baselines.\end{itemize}
%
%
\section{Problem Formulation}
\label{sec:preliminaries}
In graph theory, a graph network is defined by the interconnection between distinct nodes. A directed weighted graph is denoted as $\mathcal{G}=(V, \boldsymbol{E}, \boldsymbol{A})$, where $V$ represents the set of nodes with $\lvert V \rvert=N$, $\boldsymbol{E}$ represents the connection between these nodes and $\boldsymbol{A} \in \mathbb{R}^{N \times N} $ is the weighted adjacency matrix. The graph features within a time window $T_w$ are defined as $\boldsymbol{\mathcal{X}}=[\boldsymbol{X}_1,\boldsymbol{X}_2,\dots,\boldsymbol{X}_T] \in \mathbb{R}^{N \times d \times T_w}$, where $\boldsymbol{X}_t \in \mathbb{R}^{d \times T_w}$ is the feature matrix for the graph at the time sample $t$ and $d$ denotes the number of feature channels.  The Laplacian matrix is defined as $\boldsymbol{L}=\boldsymbol{D}-\boldsymbol{A}$, where $\boldsymbol{D} \in \mathbb{R}^{N \times N}$  is the diagonal matrix containing the degree of each node $\boldsymbol{D}_{ii}=\sum_j  \boldsymbol{A}_{(i,j)}$. The normalized Laplacian matrix is expressed as $\boldsymbol{L}=\boldsymbol{I}_N-\boldsymbol{D}^{-\frac{1}{2}}\boldsymbol{A}\boldsymbol{D}^{-\frac{1}{2}}$, where $\boldsymbol{I}_N$ is an identity matrix of order $N$. 

We validate our model on the LargeST benchmark dataset~\cite{liu2024largest}, consisting of the real-time traffic volume collected from a total of $8600$ sensor nodes, each containing five years of data from three distinct sub-regions, offers significant diversity in spatial and temporal patterns. The nodes cover the region of the Greater Bay Area (GBA: $2,352$ nodes), Greater Los Angeles (GLA: $3,834$ nodes), and San Diego (SD: $716$ nodes). This diversity tests the ability of the model to generalize across varying distributions and highlights its scalability.

In this work, we assume that the real-time data obtained from the sensors is noise-free or contains only inherent noise. If the features of the graph are corrupted by additional noise, ambient conditions may cause performance degradation. The distribution of this noise may vary spatially (with the location of the sensors) and temporally (with time intervals). We model this as   
\begin{equation}    \boldsymbol{\tilde{\mathcal{X}}}=\boldsymbol{\mathcal{X}}+\boldsymbol{\mathcal{N}}(\mu,\hat{\sigma}^2),
\label{Eq:signal_noise_model}
\end{equation}
\noindent where $\boldsymbol{\tilde{\mathcal{X}}}$ represents the noisy features of the graph obtained by adding random noise $\boldsymbol{\mathcal{N}}$ to the original graph features. The noise is modeled as independent and identically distributed (i.i.d.) Gaussian noise with mean $\mu$ and standard deviation $\hat{\sigma}$. The noisy features, along with the graph $\mathcal{G}$, are input into a forecasting framework to predict future states from time interval $(T+1)$ to $(T+N_H)$, where $N_H$ represents the future horizon length.
\begin{align*}[\boldsymbol{\tilde{X}^{(1)}},\dots,\boldsymbol{\tilde{X}^{(T)}};\boldsymbol{\mathcal{G}}] \stackrel{f}{\rightarrow}
  [\boldsymbol{X^{(T+1)}},\dots,\boldsymbol{X^{(T+N_H)}}],
\end{align*}
where $\mathit{f}$ is the function learned by the deep neural network parameterized by $\boldsymbol{\Theta}$ which can be determined by minimizing the mean absolute error (MAE), that is,
\begin{equation*}
\boldsymbol{\Theta}^*=\arg \min_{\boldsymbol{\Theta}} \;\frac{1}{N} \sum_{n=0}^{N} \lvert \mathit{f}(\boldsymbol{\Theta},\boldsymbol{\tilde{\mathcal{X}}}_n) - \boldsymbol{\mathcal{X}}_n\rvert.
\end{equation*}
\begin{figure*}[!t]
\centering
\begin{subfigure}[b]{0.8\textwidth}
\centering
    \includegraphics[width=\textwidth]{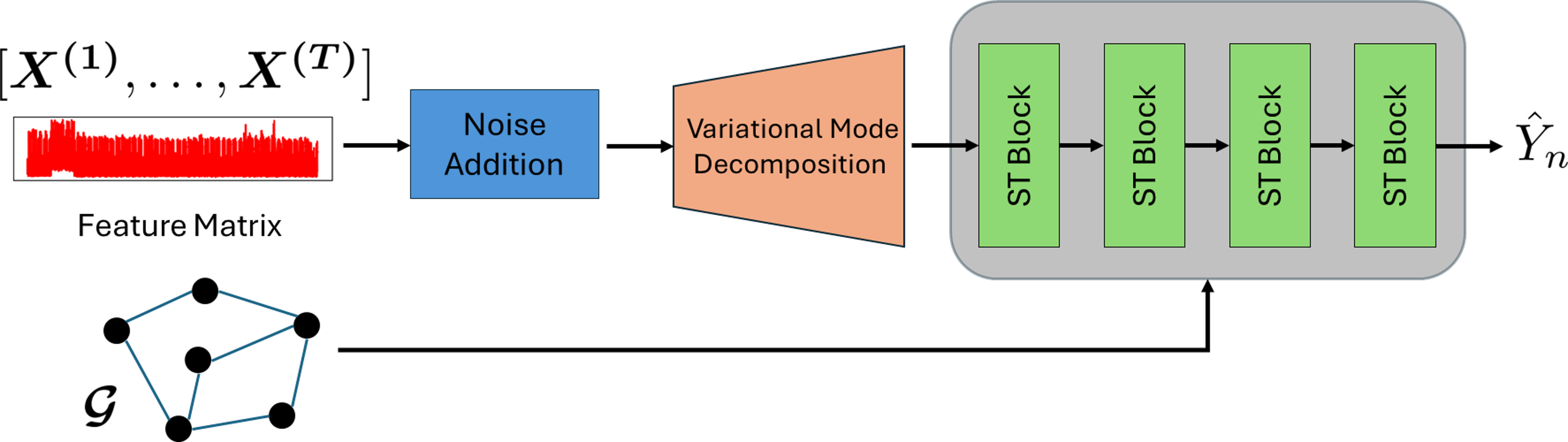}
    \caption{}
    \label{fig:block_diagram_a}
    \end{subfigure}
    \hspace{1cm}
    \begin{subfigure}[b]{0.48\textwidth}
\centering
    \includegraphics[width=\textwidth]{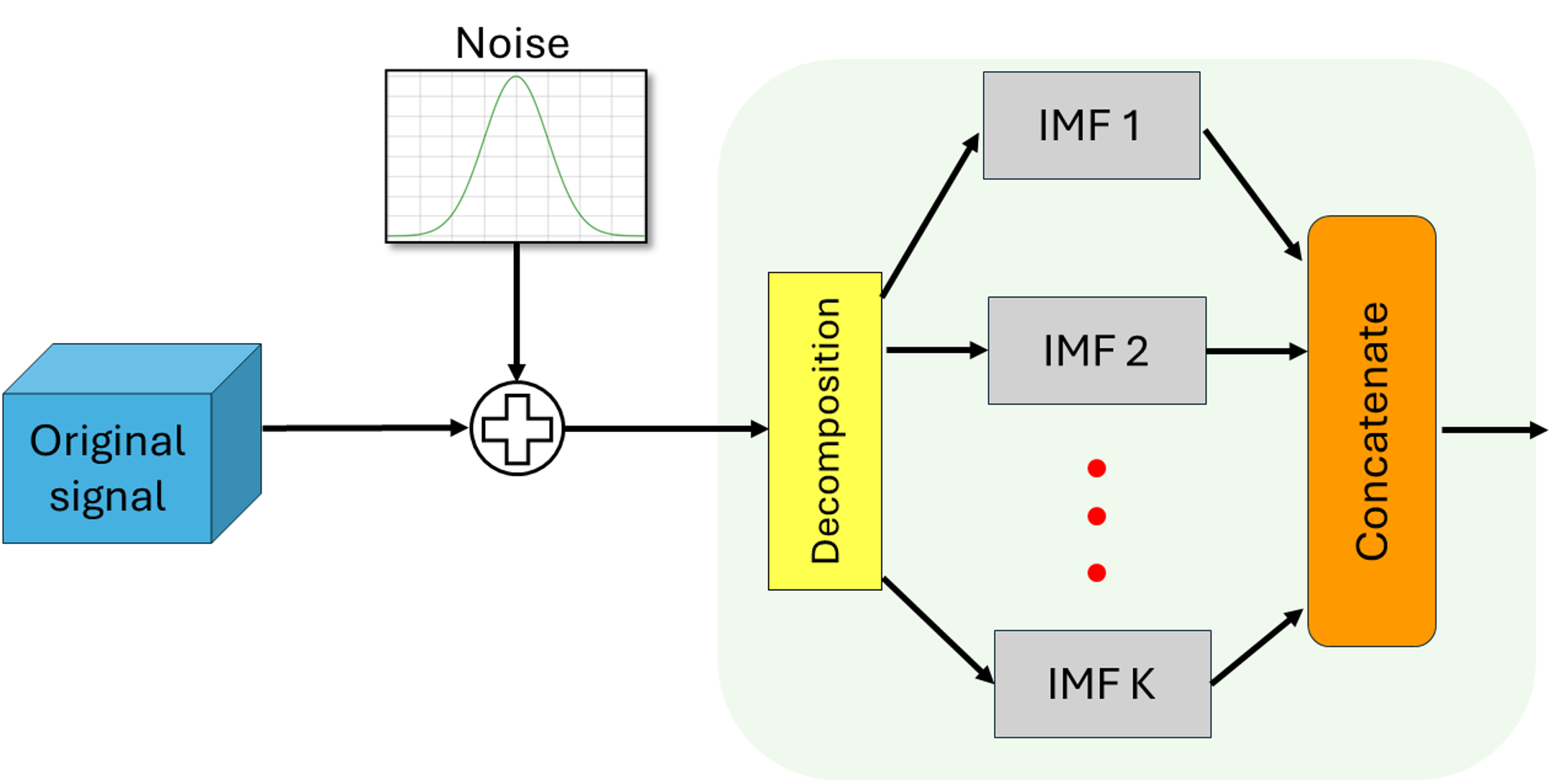}
    \caption{}
    \end{subfigure}
    \hfill
    \begin{subfigure}[b]{0.45\textwidth}
\centering
    \includegraphics[width=\textwidth]{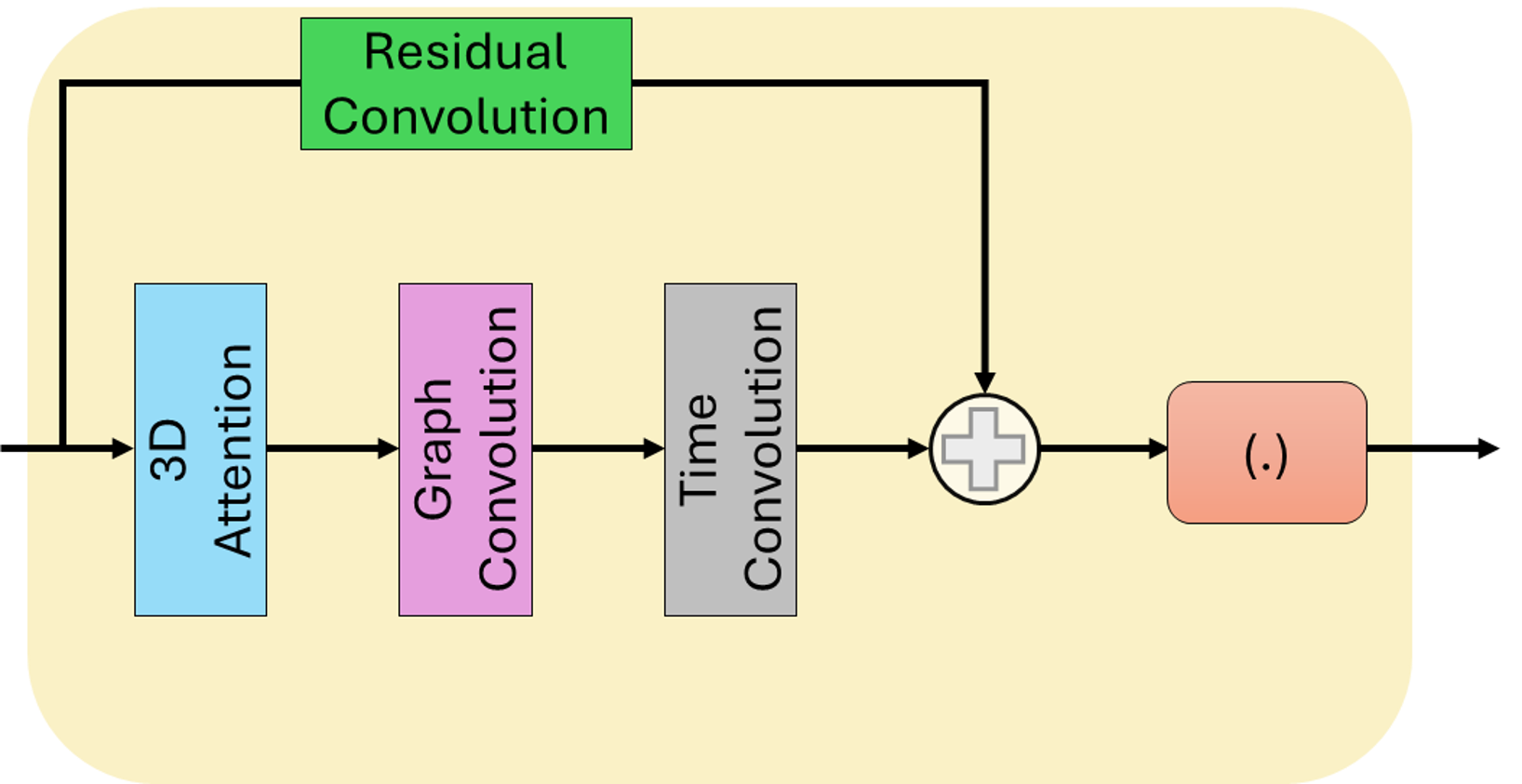}
    \caption{}
    \end{subfigure}
        \caption{(a) Proposed architecture Overview: the features of the graph $[\boldsymbol{{X}^{(1)}},\dots,\boldsymbol{{X}^{(T)}}]$ and the graph $\boldsymbol{\mathcal{G}}$ serve as an input to the pipeline and the output is the future prediction $\hat{Y}_n$. We add noise to the features of the graph to generate the noisy features $[\boldsymbol{\tilde{X}^{(1)}},\dots,\boldsymbol{\tilde{X}^{(T)}}]$ using \eqref{Eq:signal_noise_model}. The parameters of the model are determined during backpropagation using mean absolute error. (b) Variational mode decomposition: the features of the graph (signals) are decomposed into modes. (c) ST block consists of the 3-D attention, graph convolution, residual convolution, and time convolution as depicted, and (.) represents the non-linear activation function. }
    \label{fig:block_diagram}
\end{figure*}
\vspace{-1.0em}
\section{Proposed Model}
\label{sec:methodology}

\subsection{Architecture Details}
Our architecture consists of two main components, a pre-processing module and a deep neural network (DNN) block~\cite{Ahmad2024variational}, depicted in Fig.~\ref{fig:block_diagram} and summarized below:
\begin{itemize}
    \item Noise with a known distribution is added to the spatiotemporal (ST) data.
    \item Decomposition of the features of the noisy data using variational mode decomposition (VMD). 
    \item The extracted features are then processed in the DNN~\cite{Ahmad2024variational}. 
    \item Each ST block within the DNN consists of attention blocks, Chebyshev graph convolution, and a convolutional neural network (CNN). 
\end{itemize}
First, VMD maps the 1-D time series signal into a higher-dimensional representation $Z\in \mathbb{R}^{T \times K}$, where $K$ represents the number of decomposed modes, and $T$ is the length of a signal. For $N$ number of nodes in the graph, it is represented by $\boldsymbol{\mathcal{Z}}\in \mathbb{R}^{N\times T \times K}$. The primary objective of these modes is to discriminate between valuable information for forecasting and noise distribution in complex and non-stationary data. Since the higher frequency modes typically represent the noise in the data, it can be challenging to differentiate between these modes. Therefore, we introduce the channel attention within the ST block to emphasize the significance of each mode through a learning mechanism. The feature attention matrix is employed during Chebyshev graph convolution, while spatial and temporal attention mechanisms learn the hidden complex spatio-temporal representation in the data. To determine the temporal relationships, we utilize a temporal convolution network (TCN), which is based on 2-D convolution. For training the neural network, we utilize the mean absolute error (MAE) given by
\begin{equation}
    \mathcal{L} =\frac{1}{N} \sum_{n=1}^{N} |Y_n - \hat{Y}_n|,
\end{equation}
where $Y_n$ is the ground truth, and $\hat{Y}_n$ is the predicted output from the model. We employ the backpropagation algorithm to train the model until the loss function $\mathcal{L}$ converges.  
\subsection{Segregation using VMD}
Dragomiretskiy et al. proposed a method to decompose the signal based on bandwidth known as variational mode decomposition (VMD)~\cite{dragomiretskiy2013variational}. This method is commonly used for feature extraction, as seen in~\cite{Ahmad2024variational}, where each feature is referred to as an intrinsic mode function (IMF). This decomposition requires the minimization of the objective function $\sum_k\lVert\partial_t[(\delta(t)+\frac{j}{\pi t})*u_k(t)] e^{-j\omega_kt}\rVert_2^2$ under the reconstruction constraint $\sum_k u_k=f$, where `$*$' represents the convolution operation, $u_k(t)$ is the mode of the signal, and $\omega_k$ is center frequency of the $k^{\rm {th}}$ mode.
This minimization problem is solved using the Lagrangian method, and the convex problem is solved via the alternating direction method of multipliers (ADMM). The formulations for calculating the mode and center frequency in the frequency domain are given by
\begin{equation}
\label{eq:mode}
   \hat{u}_k^{(n+1)}=\frac{\hat{f}(\omega)-\sum\limits_{i< k} \hat{u_i}^{n+1}(\omega)-\sum\limits_{i > k}\hat{u_i}^{n}(\omega)+\frac{\hat{\lambda}(\omega)}{2}}{1+2\alpha(\omega-\omega^n_k)^2},
\end{equation}
\begin{equation}
\label{eq:center_frequency}
\omega_k^{n+1}=\frac{\displaystyle \sum\limits_{\omega=T}^{2T} \omega |\hat{u}_k^{n+1}(\omega)|^2d\omega}{\displaystyle \sum\limits_{\omega=T}^{2T} |\hat{u}_k^{n+1}(\omega)|^2d\omega},
\end{equation}
\noindent where $\lambda$ is the Lagrangian multiplier, and $\alpha$ is the bandwidth constraint. To prevent discontinuity at the signal boundaries, some preprocessing involves applying the mirror operation that corresponds to Neumann boundary conditions. We primarily implement \eqref{eq:mode}  and \eqref{eq:center_frequency} to determine the components of ST signals. 
\subsection{3D Attention}
In spatio-temporal forecasting, capturing spatial and temporal correlations is essential. Guo et al.~\cite{guo2019attention} employed spatial and temporal attention mechanisms to enhance traffic flow forecasting. Shi et al.~\cite{shi2020spatial} utilized an encoder-decoder framework to model spatial and temporal dependencies separately. Ding et al.~\cite{ding2020interpretable} proposed binary or weighted spatiotemporal attention mechanisms to more effectively capture spatial and temporal dependencies in time-series data. Channel attention has also been integrated into convolutional neural networks (CNNs) to boost performance~\cite{wang2020eca}. Yan et al. combined CNNs with spatial-channel and temporal attention mechanisms to predict future grasp stability~\cite{yan2021sct}. More recently, Nie et al.~\cite{nie2024triplet} introduced a triplet self-attention mechanism (spatial, temporal, and channel) for spatio-temporal predictive models.

\begin{figure}[!tb]
\centering
\begin{subfigure}[b]{0.8\linewidth}
     \centering
    \includegraphics[width=\linewidth]{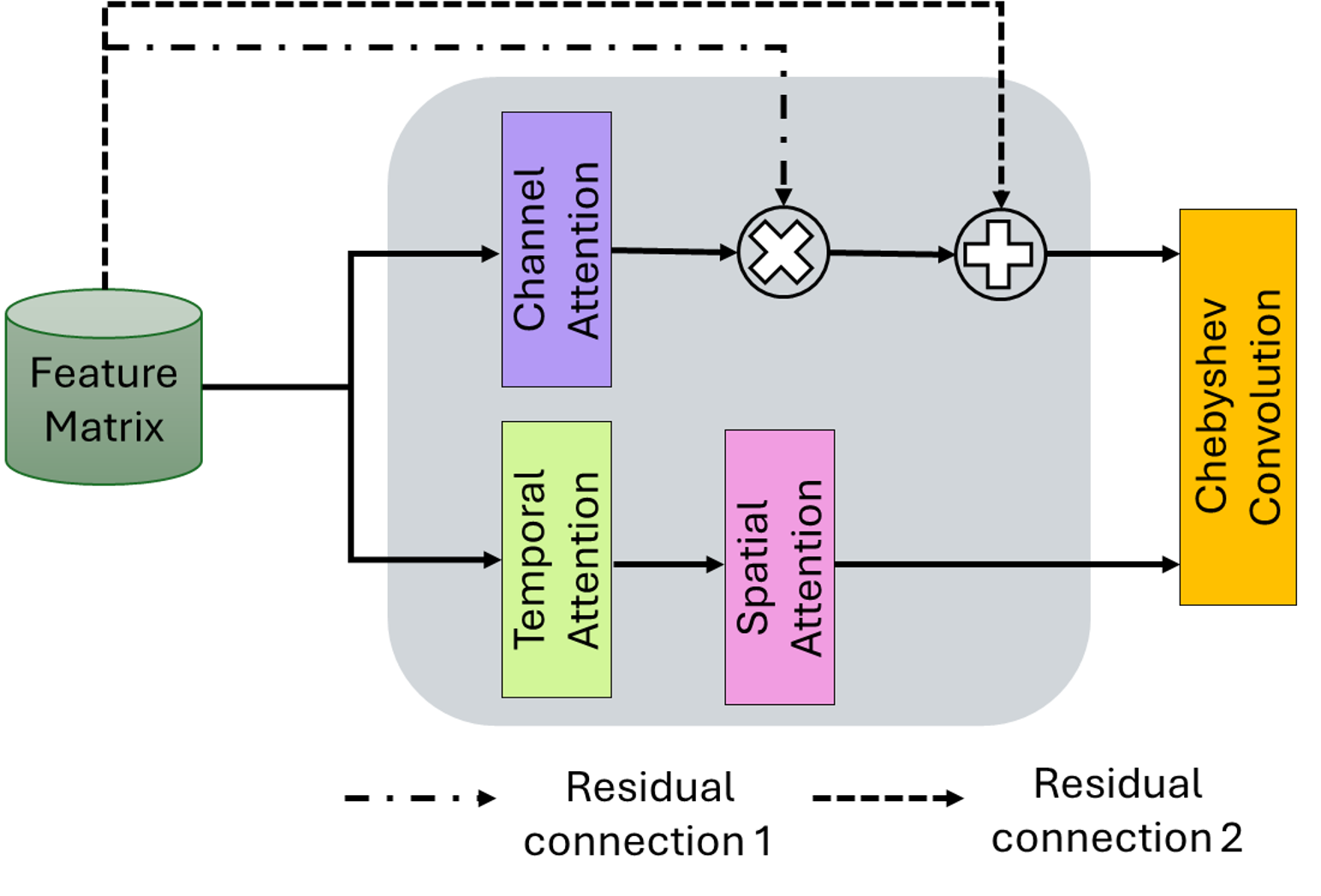}
    \caption{}
    \label{fig:3d_attention}
\vspace{1.0em}
\end{subfigure}
\centering
   \begin{subfigure}[b]{0.8\linewidth}
       \centering
    \includegraphics[width=\linewidth]{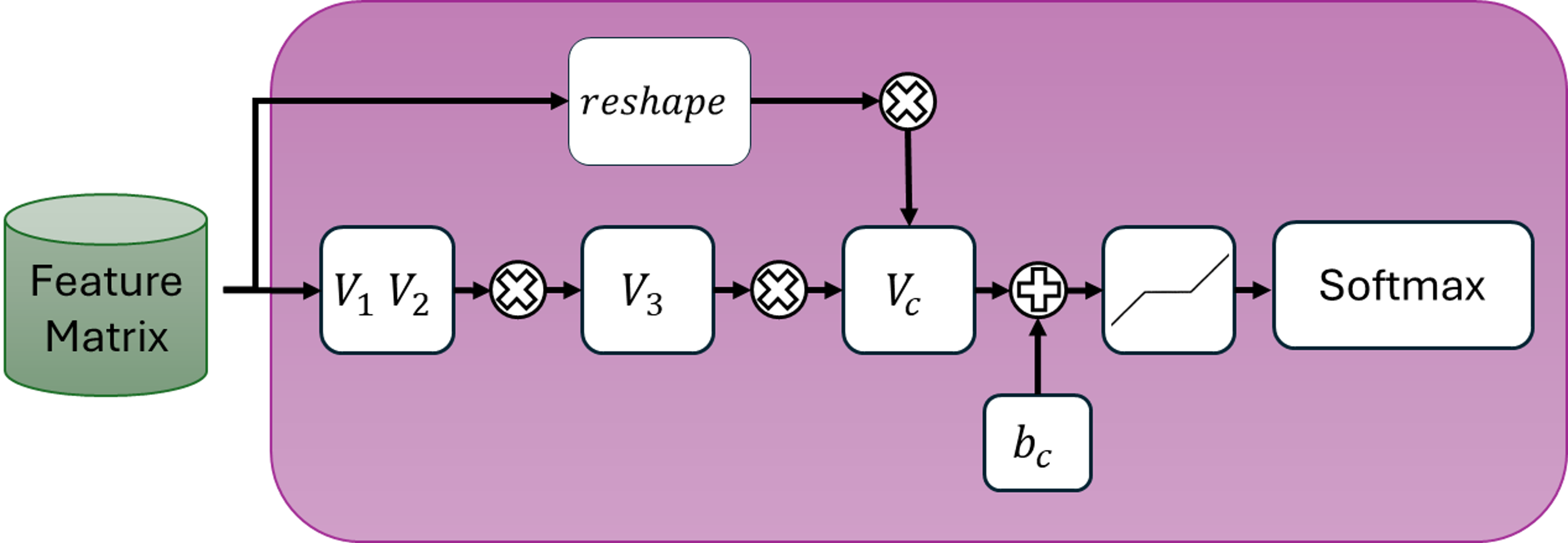}
    \caption{}
    \label{fig:channel_attention}
   \end{subfigure}
    \caption{(a) 3D attention mechanism consists of spatial, temporal, and channel components. The channel and temporal attentions are computed in parallel, while the normalized temporal attention matrix serves as an input to the spatial attention. Residual connection 1 imposes attention on the feature matrix using a multiplication operator, and residual connection 2 further adds the noisy feature to the attention-enhanced feature matrix. (b) Channel attention: $V_1, V_2, V_3$ represent the projection trainable weights and soft-thresholding is applied before the softmax activation function.}
    \label{fig:attention}
\end{figure}

In this work, we employ a 3D attention mechanism with dynamic relation extraction~\cite{feng2017effective} to model complex spatial and temporal dependencies. The 3D channel attention mechanism leverages learnable thresholds to suppress irrelevant or noisy decomposed components. This ensures that only significant features contribute to forecasting. This enhances the signal-to-noise ratio by dynamically adapting to the underlying data distribution. Fig.~\ref{fig:3d_attention} illustrates the flow diagram of the 3D attention mechanism.
\subsubsection{Spatial Attention}
Spatial attention determines the relationships among nodes in the graph and is expressed as
\begin{equation}
\mathbf{S}=\mathbf{V}_s \; \sigma((\boldsymbol{\mathcal{Z}}\mathbf{W}_1\mathbf{W}_2(\mathbf{W}_3\boldsymbol{\mathcal{Z}})^\textit{T})+\mathbf{b}_s),
\end{equation}
\begin{equation}
\label{eq:normalizd_spatial}
\mathbf{S}^{'}_{i,j}=\frac{\exp(\mathbf{S}_{i,j})}{\sum\limits_{j=1}^{N} \exp(\mathbf{S}_{i,j})},
\end{equation}
where $\mathbf{V}_s,\: \mathbf{b}_s \in \mathbb{R}^{N \times N}, \:\mathbf{W}_1\in \mathbb{R}^{T_w},\: \mathbf{W}_2 \in \mathbb{R}^{d \times T_w}, \: \mathbf{W}_3 \in \mathbb{R}^{d}$ are trainable parameters to capture the spatial correlation, and $\sigma$ is the sigmoid activation function applied to each element of the matrix. $\mathbf{S}^{'}_{i,j}$ is the normalized spatial correlation matrix between node $i$ and node $j$.
\subsubsection{Temporal Attention} 
To dynamically capture the correlation across time, temporal attention is applied as
\begin{equation}
\mathbf{E}=\mathbf{V}_e \; \sigma(((\boldsymbol{\mathcal{Z}}^\textit{T}\mathbf{U}_1)\mathbf{U}_2(\mathbf{U}_3\boldsymbol{\mathcal{Z}}))+\mathbf{b}_e),
\end{equation}
\begin{equation}
\label{eq:normalized_temporal}
\mathbf{E}^{'}_{i,j}=\frac{\exp(\mathbf{E}_{i,j})}{\sum\limits_{j=1}^{T_w} \exp(\mathbf{E}_{i,j})},
\end{equation}
where $\mathbf{V}_e,\: \mathbf{b}_e \in \mathbb{R}^{T_w \times T_w}, \:\mathbf{U}_1 \in \mathbb{R}^{N},\: \mathbf{U}_2 \in \mathbb{R}^{N \times d}, \: \mathbf{U}_3 \in \mathbb{R}^{d}$ are trainable parameters to express the temporal attention. $\mathbf{E}^{'}_{i,j}$ is the normalized temporal correlation matrix between node $i$ and node $j$.
\subsubsection{Channel Attention}
To differentiate between low-frequency modes and high-frequency modes, we introduce channel attention in our proposed architecture. High-frequency modes typically contain noise, while low-frequency modes represent the overall signal trend. To highlight the significance of each mode on prediction performance, the learnable attention mechanism is elaborated in Fig.~\ref{fig:channel_attention}. 
\begin{equation}
\mathbf{C}=\mathbf{V}_c \; \sigma((\boldsymbol{\mathcal{Z}}\mathbf{V}_1)\mathbf{V}_2(\mathbf{V}_3\boldsymbol{\hat{\mathcal{Z}}})+\mathbf{b}_c),
\end{equation}
\noindent where $\mathbf{V}_c,\: \mathbf{b}_c \in \mathbb{R}^{d \times d}, \:\mathbf{V}_1\in \mathbb{R}^{T_w},\: \mathbf{V}_2 \in \mathbb{R}^{N \times T_w}, \: \mathbf{V}_3 \in \mathbb{R}^{N}$ are trainable weights that determine the channel attention. The feature matrix is reshaped as $\boldsymbol{\mathcal{Z}} \in \mathbb{R}^{d \times N \times T_w}$ and $\boldsymbol{\hat{\mathcal{Z}}} \in \mathbb{R}^{N \times T_w \times d}$. The element-wise soft thresholding shrinkage operator is defined as~\cite{beck2009fast} : 
\begin{equation}
    \mathbf{C}_{\rm{Th}} =\rm{sign}(\mathbf{C}) \rm{max}(\lvert \mathbf{C} \rvert- \Phi,0),
\end{equation}
\begin{equation}
\mathbf{C}^{'}_{i,j}=\frac{\exp(\mathbf{C}_{\rm {Th}(i,j)})}{\sum\limits_{j=1}^{N} \exp(\mathbf{C}_{\rm {Th}(i,j)})},
\end{equation}
\noindent where $\Phi$ is a learnable thresholding parameter, and $\mathbf{C}_{\rm{Th}}$ is an attention matrix. The normalized channel attention on the features matrix to determine the feature attention matrix $\boldsymbol{\mathcal{Z}}_{\rm{att}}$ can be obtained as 
\begin{equation}
\label{eq:feature_att}
   z_{\rm {att}{(i,j)}}^{(t)} =\sum_{k=1}^{d} z_{i,k}^{(t)}  \mathbf{C}^{'}_{k,j},
\end{equation}
\begin{equation}
\label{eq:residual_feature}
\boldsymbol{\mathcal{Z}}_{\rm {new}}=\boldsymbol{\mathcal{Z}}_{\rm{att}}+\boldsymbol{\mathcal{Z}}.
\end{equation}
A residual connection is added to extract the better feature representations, as shown in \eqref{eq:residual_feature}. The updated $\boldsymbol{\mathcal{Z}}_{\rm{new}}$ is then used in temporal convolution, residual convolution, and graph convolution.  
\subsection{Spectral Graph Filtering}
Spectral graph filtering is a method used to process graph signals, primarily for smoothing and denoising applications. Here, the polynomial filters are treated as learnable parameters. Examples of such methods include ChebNet~\cite{kipf2016semi}, CayleyNet~\cite{levie2018cayleynets}, Monomial polynomials~\cite{chien2020adaptive}, and Bernstein polynomials~\cite{he2021bernnet}. The formulation of ChebNet utilized in our work is given by
\begin{equation}
    g_\theta(\mathbf{L})=\sum_{m=0}^{M-1} \theta_m (T_m({\mathbf{\hat L}})\odot \mathbf{S}^{'})\boldsymbol{\mathcal{Z}}_{\rm {new}},
\end{equation}
where $\theta_m$ represents the Chebyshev coefficient of order $m$, and $T_m({\mathbf{\hat L}})$ is the Chebyshev polynomial of order $m$. Here, $\hat{\mathbf{L}}=\frac{2}{\lambda_{\rm {max}}}\mathbf{L}-I_n$, with $\lambda_{\rm {max}}$ being the maximum eigenvalue of the Laplacian matrix. The graph convolutional kernel $g_\theta$ is solved to extracting the information from $0$ to $(M-1)^{\rm{th}}$ order.
\section{Experiments}
\label{sec:results}
\subsection{Evaluation Metrics}
 We assess the performance of the models using three commonly known regression metrics: mean absolute percentage error (MAPE), mean absolute error (MAE), and root mean square error (RMSE). These metrics allow us to comprehensively compare our proposed network with baseline models. 

The experiments are conducted on a Linux system with an Intel i9 processor, 24 GB RAM, and an NVIDIA 3080Ti GPU. The model is trained using 2019 data sampled at 15-minute intervals. The data is split into 60\% training, 20\% validation, and 20\% testing. The batch size is set to $48$ for the SD region and  $4$ for the GLA and GBA regions. The historical input size is $12$, and we predict the next $12$ horizon states, corresponding to a long-term traffic prediction for the next $3$ hours. The proposed methods are implemented using PyTorch, and the VMD decomposition is performed using the vmdpy tool~\cite{carvalho2020evaluating}. The model is trained using the Adam optimizer for $100$ epochs with early stopping criteria.
\subsection{Baseline Models}
In this work, we compare the performance of our proposed model with the following baseline networks: historical last (HL)~\cite{liang2021revisiting}, long short-term memory (LSTM)~\cite{Hochreiter1997long}, diffusion convolutional recurrent neural network (DCRNN)~\cite{li2017diffusion}, adaptive graph convolutional recurrent network (AGCRN)~\cite{bai2020adaptive}, graph WaveNet (GWNET)~\cite{wu2019graph}, spatial-temporal graph ODE networks (STGODE)~\cite{fang2021spatial}, dynamic spatial-temporal aware graph neural network (DSTAGNN)~\cite{lan2022dstagnn}, dynamic graph convolutional recurrent network (DGCRN)~\cite{li2023dynamic}, decoupled dynamic spatial-temporal graph neural network (D$^2$STAGNN)~\cite{shao2022decoupled}, attention-based spatial-temporal graph convolutional network (ASTGCN)~\cite{guo2019attention}, and variational mode graph convolutional network (VMGCN)~\cite{Ahmad2024variational}. 
\begin{figure}[!t]
\begin{subfigure}[b]{0.485\linewidth}
    \centering
    \includegraphics[width=\linewidth]{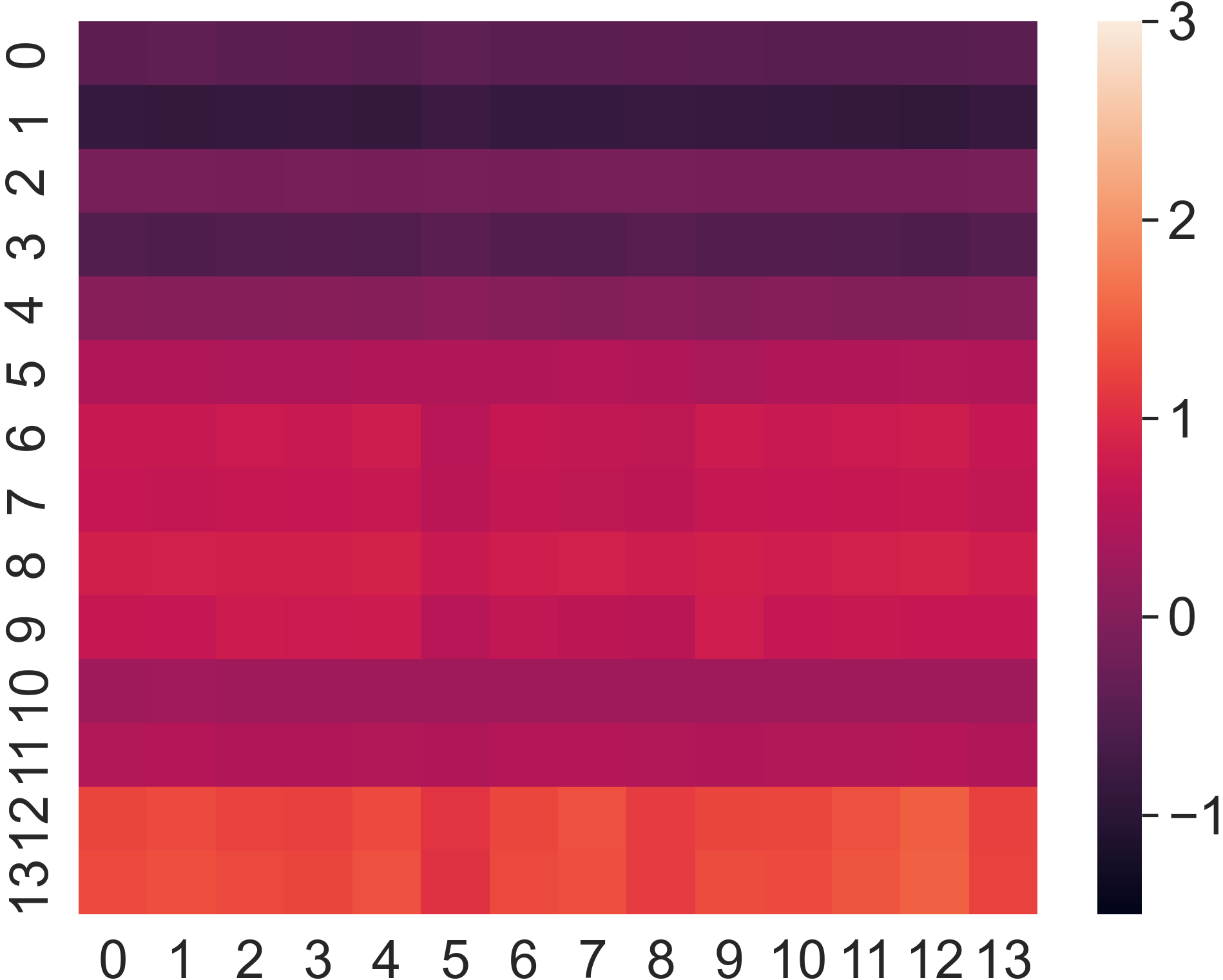}
    \caption{}
    \label{fig:GBA_noise_free}
    \end{subfigure}
    \hfill
\begin{subfigure}[b]{0.485\linewidth}
    \centering
    \includegraphics[width=\linewidth]{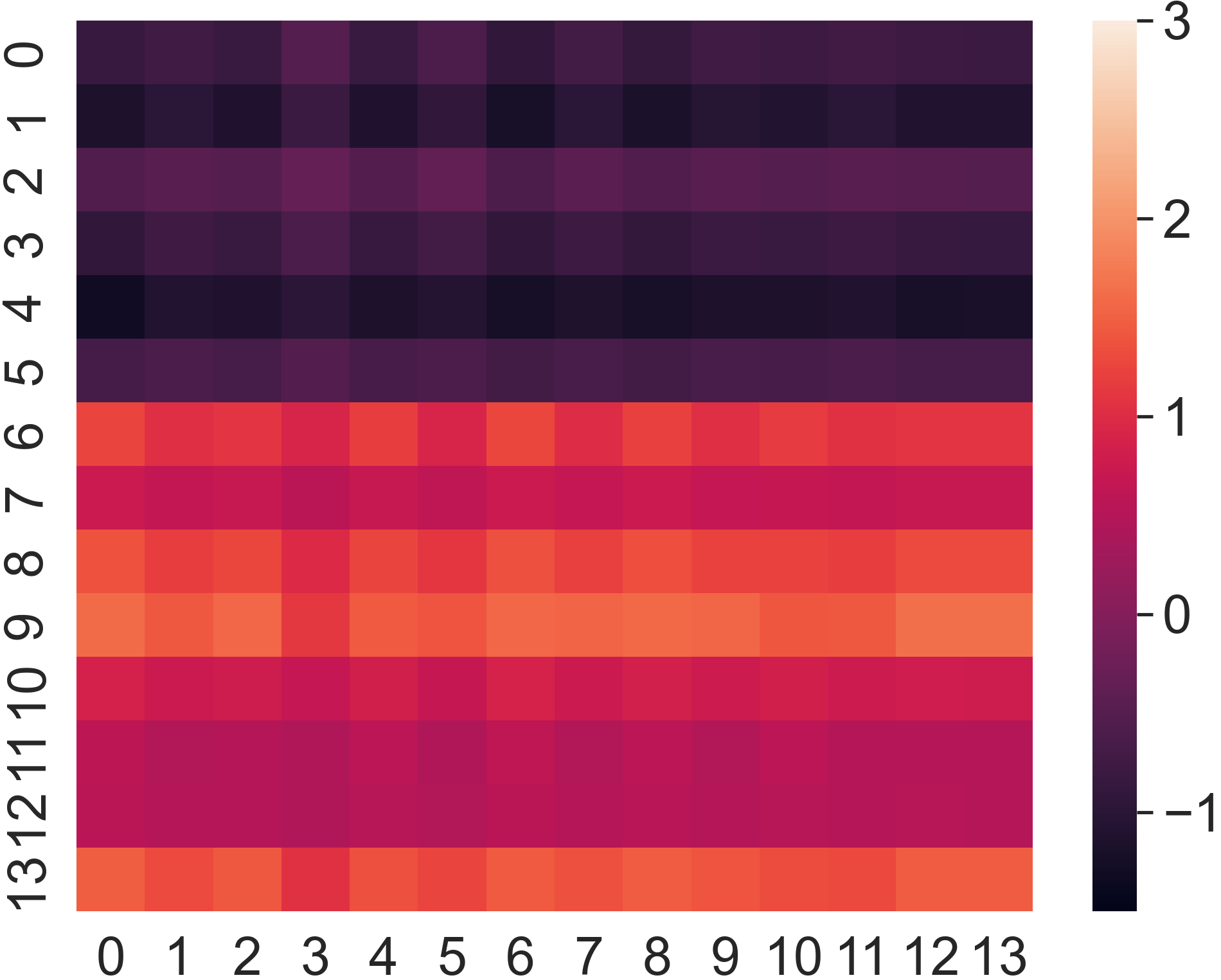}
    \caption{}
    \label{fig:GBA_noise}
    \end{subfigure}
\begin{subfigure}[b]{0.485\linewidth}
    \centering
    \includegraphics[width=\linewidth]{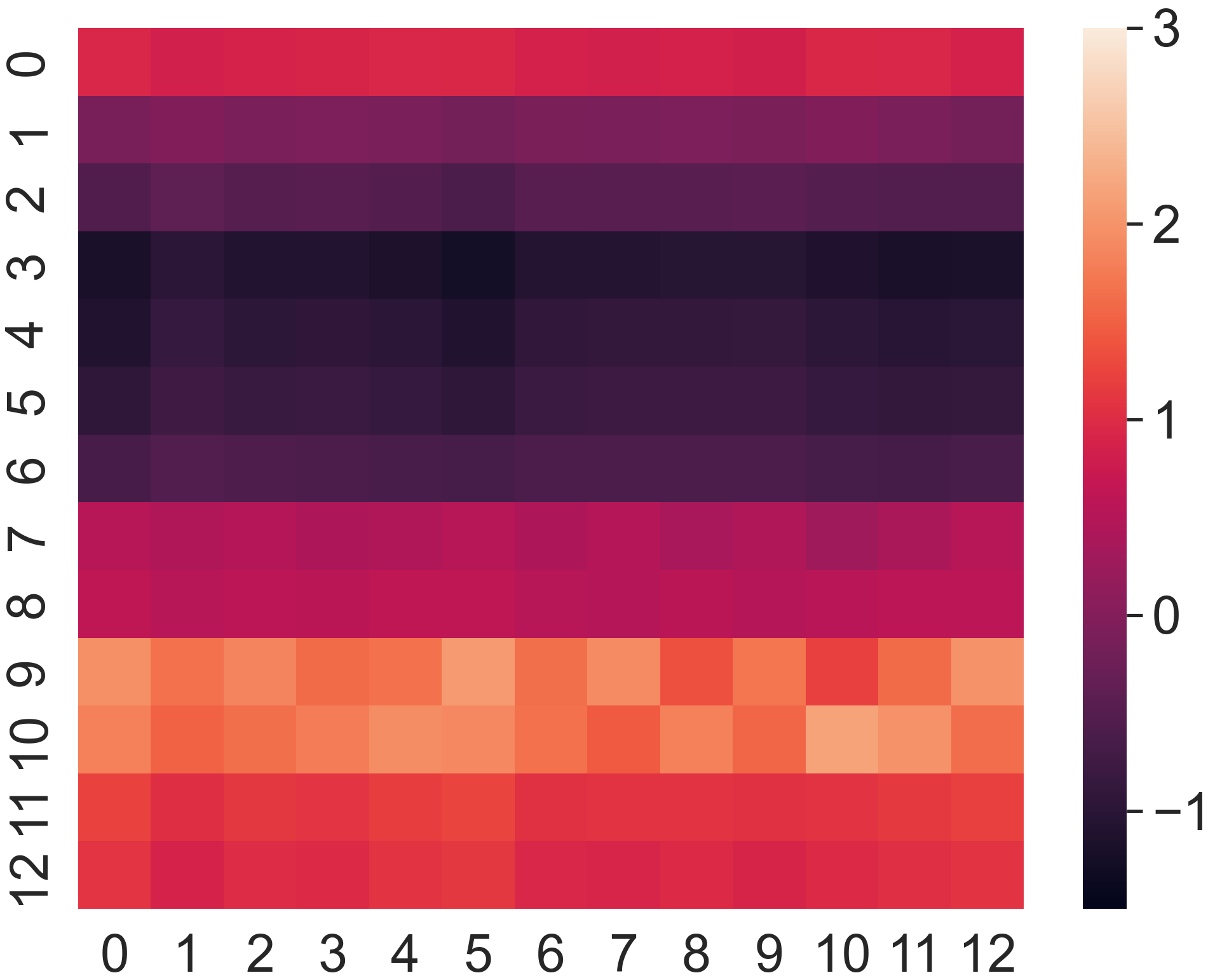}
    \caption{}
    \label{fig:GLA_noise_free}
    \end{subfigure}
    \hfill
    \begin{subfigure}[b]{0.485\linewidth}
    \centering
    \includegraphics[width=\linewidth]{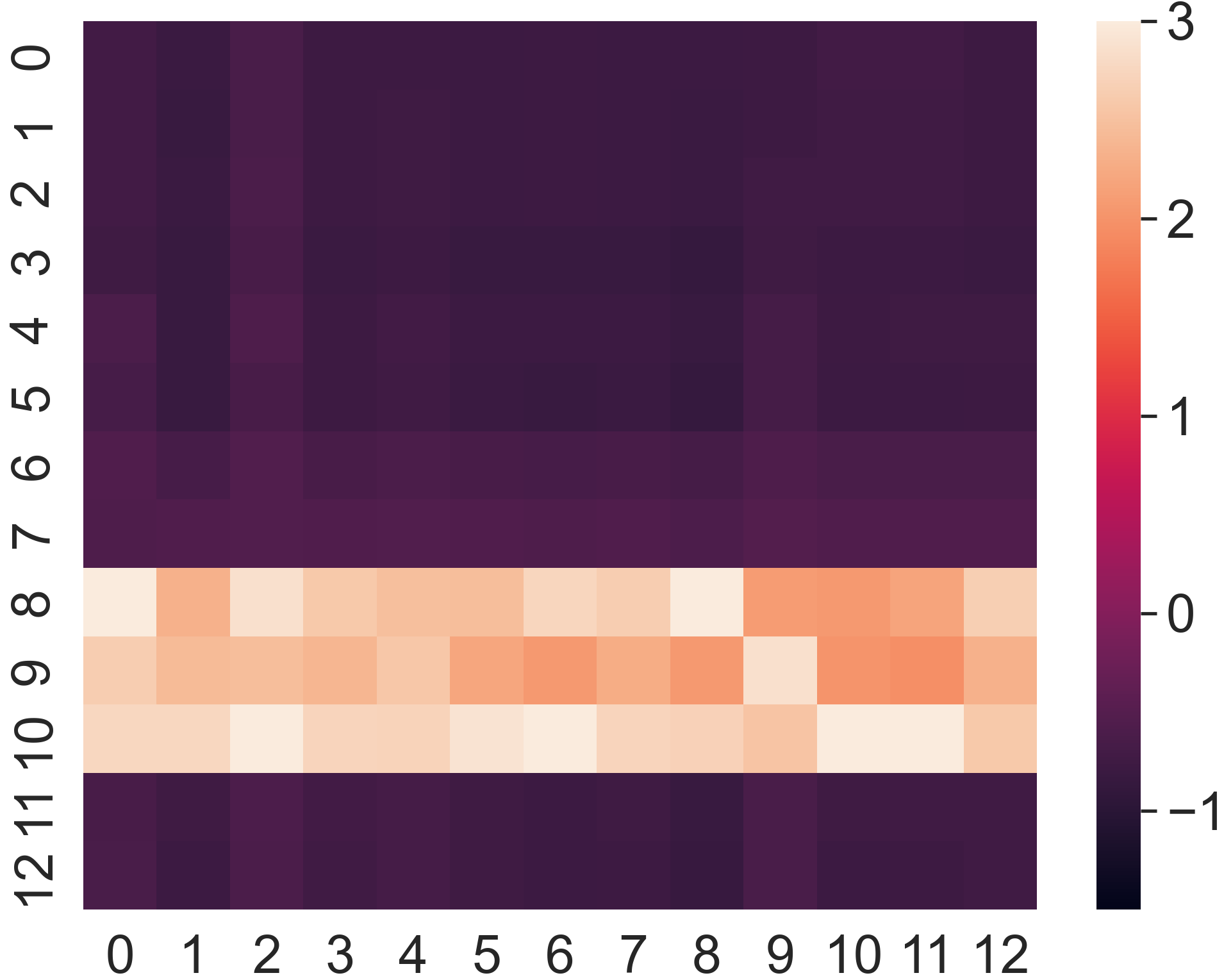}
    \caption{}
    \label{fig:GLA_noise}
    \end{subfigure}
    \caption{Thresholded channel matrix $C_{\rm{TH}}$. For $K=14$ and GBA region, (a) $\hat{\sigma}=0$ and (b) $\hat{\sigma}=0.1$. For $K=13$ and GLA region, (c) $\hat{\sigma}=0$ and (d) $\hat{\sigma}=0.1$.}
    \label{fig:heatmap}
\end{figure}
\subsection{Performance Study}
We analyze the performance of the model under noise deviation of $\hat{\sigma}=0$ and $\hat{\sigma}=0.1$. The number of modes utilized for our simulations is selected $13$ for SD, GLA, and $14$ for GBA (Ahmad et al. 2024). A significant increase in amplitude is observed in higher-frequency modes relative to the low-frequency modes in the presence of additive noise. In Fig.~\ref{fig:heatmap}, we plot the thresholded channel attention matrix $C_{\rm{TH}}$ for GBA and GLA regions, where it can be observed that these matrices have positive values for high-frequency modes and negative values for low-frequency modes. The transition of thresholded matrices from $\hat{\sigma}=0$ to $\hat{\sigma}=0.1$ indicates that more low-frequency components shifted towards the negative values. The main purpose of this attention is to suppress the noisy components that lead to better prediction accuracy.

\begin{table*}[!t]
\centering
\caption{Comparison of performance evaluation metrics MAE, MAPE, and RMSE between different baseline and our model on horizon $3$, $6$, $12$, and average. The average results are computed using the mean from the horizon of $1$ to $12$. The number of parameters (param) is described in K (kilo), $10^3$ and M (million), $10^6$, and the best performance metrics are highlighted in red bold numbers. In contrast, the numbers in blue bold indicate the best result for $\hat{\sigma}=0.1$. \Cross 
 \: indicates the results computed for $\hat{\sigma}=0.0$. * numbers are taken from~\cite{liu2024largest} and ** numbers are taken from~\cite{Ahmad2024variational}.
 }
\resizebox{.93\textwidth}{!}{
\begin{threeparttable}
\begin{tabular}{c|cc|ccc|ccc|ccc|ccc}

\toprule
 Dataset & Method & Param & \multicolumn{3}{c}{Horizon 3} & \multicolumn{3}{c}{Horizon 6} & \multicolumn{3}{c}{Horizon 12} & \multicolumn{3}{c}{Average}  \\
\cmidrule(r){4-15}
& & &MAE & RMSE & MAPE& MAE & RMSE& MAPE & MAE & RMSE & MAPE& MAE & RMSE & MAPE \\
\midrule
\multirow{13}{*}{GBA} 
& HL\Cross* & - & 32.57 & 48.42 & 22.78\% & 53.79 & 77.08 & 43.01\% & 92.64 & 126.22 & 92.85\% & 56.44 & 79.82 & 48.87\% \\
& LSTM\Cross* & 98K & 20.41 & 33.47 & 15.60\% & 27.50 & 43.64 & 23.25\% & 38.85 & 60.46 & 37.47\% & 27.88 & 44.23 & 24.31\% \\
& DCRNN\Cross* & 373K & 18.25 & 29.73 & 14.37\% & 22.25 & 35.04 & 19.82\% & 28.68 & 44.39 & 28.69\% & 22.35 & 35.26 & 20.15\% \\
& AGCRN\Cross* & 777K & 18.11 & 30.19 & 13.64\% & 20.86 & 34.42 & 16.24\% & 24.06 & 39.47 & 19.29\% & 20.55 & 33.91 & 16.06\% \\
& STGCN\Cross* & 1.3M & 20.62 & 33.81 & 15.84\% & 23.19 & 37.96 & 18.09\% & 26.53 & 43.88 & 21.77\% & 23.03 & 37.82 & 18.20\% \\
& GWNET\Cross* & 344K & 17.74 & 28.92 & 14.37\% & 20.98 & 33.50 & 17.77\% & 25.39 & 40.30 & 22.99\% & 20.78 & 33.32 & 17.76\% \\
& STGODE\Cross* & 788K & 18.80 & 30.53 & 15.67\% & 22.19 & 35.91 & 18.54\% & 26.27 & 43.07 & 22.71\% & 21.86 & 35.57 & 17.76\% \\
& DSTAGNN\Cross* & 26.9M  & 19.87 & 31.54 & 16.85\% & 23.89 & 38.11 & 19.53\% & 28.48 & 44.65 & 24.65\% & 23.39 & 37.07 & 19.58\% \\
& DGCRN\Cross* & 374K & 18.09 & 29.27 & 15.32\% & 21.18 & 33.78 & 18.59\% & 25.73 & 40.88 & 23.67\% & 21.10 & 33.76 & 18.58\% \\
& D$^2$STGNN\Cross* & 446K & 17.20 & 28.50 & 12.22\% & 20.80 & 33.53 & 15.32\% & 25.72& 40.90 & 19.90\% & 20.71 & 33.44 & 15.23\% \\
& ASTGCN\Cross* & 22.30M & 21.40 & 33.61 & 17.65\% & 26.70 & 40.75 & 24.02\% & 33.64 & 51.21 & 31.15\% & 26.15 & 40.25 & 23.29\% \\
 & VMGCN\Cross & 22.38M &\textbf{\textcolor{red}{2.90}} & \textbf{\textcolor{red}{5.32}} & \textbf{\textcolor{red}{3.27\%}} & \textbf{\textcolor{red}{6.47}} & 11.62 & \textbf{\textcolor{red}{6.86\%}} & 16.42 & 26.45 & 17.55\% & 8.04 & 13.55 & 8.57\% \\
&Ours\Cross & 22.40M & 3.50 & 6.19 & 3.91\% & 6.59 & \textbf{\textcolor{red}{11.50}} & 6.89\% & \textbf{\textcolor{red}{14.77}} & \textbf{\textcolor{red}{23.47}} & \textbf{\textcolor{red}{15.27\%}} & \textbf{\textcolor{red}{7.77}} & \textbf{\textcolor{red}{12.90}} & \textbf{\textcolor{red}{8.14\%}} \\
\cmidrule(r){2-15}
&VMGCN$(\hat{\sigma}=0.1)$ & 22.38M & 10.28 & \textbf{\textcolor{blue}{13.70}}& 11.87\% & 11.92 & \textbf{\textcolor{blue}{16.60}} & 13.91\% & 18.56 & 27.67 & 20.44\% & 13.09 & \textbf{\textcolor{blue}{18.49}} & 14.83\% \\
&Ours$(\hat{\sigma}=0.1)$ & 22.40M & \textbf{\textcolor{blue}{9.90}} & 13.83 & \textbf{\textcolor{blue}{11.01\%}} & \textbf{\textcolor{blue}{11.85}} & 17.00 & \textbf{\textcolor{blue}{12.28\%}} & \textbf{\textcolor{blue}{15.50}} & \textbf{\textcolor{blue}{27.40}} & \textbf{\textcolor{blue}{16.31\%}} & \textbf{\textcolor{blue}{11.81}} & 18.92 & \textbf{\textcolor{blue}{12.82\%}} \\
\hline
\multirow{13}{*}{GLA} 
& HL\Cross*& - & 33.66 & 50.91 & 19.16\% & 56.88 & 83.54 & 34.85\% & 98.45 & 137.52 & 71.14\% & 56.58 & 86.19 & 38.76\% \\
& LSTM\Cross* & 98K & 20.09 & 32.41 & 11.82\% & 27.80 & 44.10 & 16.52\% & 39.61 & 61.57 & 25.63\% & 28.12 & 44.40 & 17.31\% \\
& DCRNN\Cross* & 373K & 18.33 & 29.13 & 10.78\% & 22.70 & 35.55 & 13.74\% & 29.45 & 45.88 & 18.87\% & 22.73 & 35.65 & 13.97\% \\
& AGCRN\Cross* & 792K & 17.57 & 30.83 & 10.86\% & 20.79 & 36.09 & 13.11\% & 25.01 & 44.82 & 16.11\% & 20.61 & 36.23 & 12.99\% \\
& STGCN\Cross* & 2.1M & 19.87 & 34.01 & 12.58\% & 22.54 & 38.57 & 13.94\% & 26.48 & 45.61 & 16.92\% & 22.48 & 38.55 & 14.15\% \\
& GWNET\Cross* & 374K & 17.30 & 27.72 & 10.69\% & 21.22 & 33.64 & 13.48\% & 27.25 & 43.03 & 18.49\% & 21.23 & 33.68 & 13.72\% \\
& STGODE\Cross* & 841K & 18.46 & 30.05 & 11.94\% & 22.24 & 36.68 & 14.67\% & 27.14 & 45.38 & 19.12\% & 22.02 & 36.34 & 14.93\% \\
& DSTAGNN\Cross* & 66.3M  & 19.35 & 30.55 & 11.33\% & 24.22 & 38.19 & 15.90\% & 230.32 & 48.37 & 23.51\% & 23.87 & 37.88 & 15.36\% \\
& DGCRN\Cross* & 432K & 17.63 & 8.12 & 10.50\% & 21.15 & 33.70 & 13.06\% & 26.18 & 42.16 & 17.40\% & 21.02 & 33.66 & 13.23\% \\
& D$^2$STGNN\Cross* & 284K & 19.31 & 30.07 & 11.82\% & 22.52 & 35.22 & 14.16\% & 27.46& 43.37 & 18.54\% & 22.35 & 35.11 & 14.37\% \\
& ASTGCN\Cross*  & 59.1M & 21.11 & 32.41 & 11.82\% & 27.80 & 44.67 & 17.79\% & 39.39 & 59.31 & 28.03\% & 28.12 & 44.40 & 18.62\% \\
& VMGCN\Cross** & 59.2M &\textbf{\textcolor{red}{3.88}} & 10.78 & \textbf{\textcolor{red}{3.99\%}} & 8.27 & 22.34 & \textbf{\textcolor{red}{7.85\%}} & 16.78 & 31.46 & 14.28\% & 9.22 & 20.69 & \textbf{\textcolor{red}{8.23\%}} \\
& Ours\Cross & 59.3M & 4.37 & \textbf{\textcolor{red}{8.99}} & 8.04\% & \textbf{\textcolor{red}{8.19}} & \textbf{\textcolor{red}{15.54}} & 7.86\% & \textbf{\textcolor{red}{15.15}} & \textbf{\textcolor{red}{24.08}} & \textbf{\textcolor{red}{12.42\%}} & \textbf{\textcolor{red}{8.85}} & \textbf{\textcolor{red}{15.53}} & 8.76\% \\
\cmidrule(r){2-15}
& VMGCN $(\hat{\sigma}=0.1)$ & 59.2M & 11.06 & 15.70 & 11.84\% & 13.10 & 20.63 & 13.88\% & 19.52 & 31.48 & 18.43\% & 14.10 & 21.94 & 14.22\% \\
& Ours $(\hat{\sigma}=0.1)$ & 59.3M & \textbf{\textcolor{blue}{10.64}} & \textbf{\textcolor{blue}{14.76}} & \textbf{\textcolor{blue}{10.36\%}} & \textbf{\textcolor{blue}{12.17}} & \textbf{\textcolor{blue}{17.46}} & \textbf{\textcolor{blue}{11.66\%}} & \textbf{\textcolor{blue}{16.72}} & \textbf{\textcolor{blue}{24.84}} & \textbf{\textcolor{blue}{14.69\%}} & \textbf{\textcolor{blue}{12.85}} & \textbf{\textcolor{blue}{18.54}} & \textbf{\textcolor{blue}{11.91\%}} \\
\hline
\multirow{13}{*}{SD} 
& HL\Cross* & - & 33.61 & 50.97 & 20.77\% & 57.80 & 84.92 & 37.73\% & 101.74 & 140.14 & 76.84\% & 60.79 & 87.40 & 41.88\% \\
& LSTM\Cross* & 98K & 19.17 & 30.75 & 11.85\% & 26.11 & 41.28 & 16.53\% & 38.06 & 59.63 & 25.07\% & 26.73 & 42.14 & 17.17\% \\
& DCRNN\Cross* & 373K & 17.01 & 27.33 & 10.96\% & 20.80 & 33.03 & 13.72\% & 26.77 & 42.49 & 18.57\% & 20.86 & 33.13 & 13.94\% \\
& AGCRN\Cross* & 761K & 16.05 & 28.78 & 11.74\% & 18.37 & 32.44 & 13.37\% & 22.12 & 40.37 & 16.63\% & 18.43 & 32.97 & 13.51\% \\
& STGCN\Cross* & 508K & 18.23 & 30.60 & 13.75\% & 20.34 & 34.42 & 15.10\% & 23.56 & 41.70 & 17.08\% & 20.35 & 34.70 & 15.13\% \\
& GWNET\Cross* & 311K & 15.49 & 25.45 & 9.90\% & 18.17 & 30.16 & 11.98\% & 22.18 & 37.82 & \textbf{\textcolor{red}{15.41\%}} & 18.12 & 30.21 & 12.08\% \\
& STGODE\Cross* & 729K & 16.76 & 27.26 & 10.95\% & 19.79 & 32.91 & 13.18\% & 23.60 & 41.32 & 16.60\% & 19.52 & 32.76 & 13.22\% \\
& DSTAGNN\Cross* & 3.9M  & 17.83 & 28.60 & 11.08\% & 21.95 & 35.37 & 14.55\% & 26.83 & 46.39 & 19.62\% & 21.52 & 35.67 & 14.52\% \\
& DGCRN\Cross* & 243K & 15.24 & 25.46 & 10.09\% & 17.66 & 29.65 & 11.77\% & 21.59 & \textbf{\textcolor{red}{35.55}} & 16.88\% & 17.38 & 28.92 & 12.43\% \\
& D$^2$STGNN\Cross** & 406K & 15.76 & 25.71 & 11.84\% & 18.81 & 30.68 & 14.39\% & 23.17 & 38.76 & 18.13\% & 18.71 & 30.77 & 13.99\% \\
& ASTGCN\Cross** & 2.15M & 19.68 & 31.53 & 12.20\% & 24.45 & 38.89 & 15.36\% & 31.52 & 49.77 & 22.15\% & 26.07 & 38.42 & 15.63\% \\
&VMGCN\Cross  & 2.17M&\textbf{\textcolor{red}{6.67}} & 13.51 & \textbf{\textcolor{red}{6.02\%}} & \textbf{\textcolor{red}{11.25}} & 27.96 & 10.23\% & 20.73 & 85.97 & 20.80\% & 12.23 & 39.48& 11.69\% \\
&Ours\Cross & 2.19M& 7.17 & \textbf{\textcolor{red}{12.27 }}& 6.08\% & 11.27 & \textbf{\textcolor{red}{20.46}} & \textbf{\textcolor{red}{9.20\%}} & \textbf{\textcolor{red}{18.44}} & 38.97 & 15.89\% & \textbf{\textcolor{red}{11.71}} & \textbf{\textcolor{red}{22.56}} & \textbf{\textcolor{red}{9.79\%}} \\
\cmidrule(r){2-15}
&VMGCN $(\hat{\sigma}=0.1)$ & 2.17M &\textbf{\textcolor{blue}{10.88}} & \textbf{\textcolor{blue}{14.92}} & 11.61\% & \textbf{\textcolor{blue}{13.15}} & \textbf{\textcolor{blue}{18.83}} & 14.11\% & \textbf{\textcolor{blue}{19.62}} & \textbf{\textcolor{blue}{28.43}} & 21.97\% & \textbf{\textcolor{blue}{14.04}} & \textbf{\textcolor{blue}{20.02}}& 15.40\% \\
&Ours$(\hat{\sigma}=0.1)$& 2.19M& 11.17 & 19.18 & \textbf{\textcolor{blue}{10.50\%}} & 13.68 & 22.47 & \textbf{\textcolor{blue}{12.03\%}} & 22.00 & 77.60 & \textbf{\textcolor{blue}{19.01\%}} & 15.04 & 37.17 & \textbf{\textcolor{blue}{13.23\%}}\\
\hline
\end{tabular}
\end{threeparttable}
}
\label{Tab:comparison}
\end{table*}
In Table~\ref{Tab:comparison}, we present a comparison of performance metrics between baseline methods and our proposed architecture across prediction horizons from 1 to 12. Horizons up to 4 are considered short-term predictions, while those beyond 4 are treated as long-term predictions. Our model demonstrates comparable performance to VMGCN, with a significant improvement in long-term predictions when additive Gaussian noise is introduced. The improvement stems from the channel attention mechanism, which increases robustness to noise by modeling its distribution and more accurately capturing future trends~\cite{li2022generative}.  

To evaluate the robustness of the proposed model, the Gaussian noise with different standard deviations $\hat{\sigma}=[0,0.1,0.2,0.5,0.8,1.0]$ is also added to the real-time traffic data. This data is decomposed and performed inference on the trained weights for $\hat{\sigma}=0.1$. The overall trend for MAE, MAPE, and RMSE shows reasonable performance on $\hat{\sigma}\pm 0.1$. It can also be observed that the short-term prediction capabilities of the model are improved on the short horizon $3$ for $\hat{\sigma}=0$.  
However, prediction performance declines as the noise deviation increases. The error metrics show a steeper increase at the intermediate horizon 6 as noise levels rise, while the slope decreases for the long-term horizon 12. This degradation in performance at higher noise levels can be attributed to lower SNR values when the temporal signals are highly contaminated with the noise.

\begin{algorithm}[!t]
\centering
\caption{Multiple-ensemble Training}
\label{alg:1}
\begin{algorithmic}
\State \textbf{Input:} Data $\boldsymbol{\mathcal{X}}$, number of epochs, noise level $\hat{\sigma}$, initial model parameters $\Theta$
\State \textbf{Initialize:} $l$ (number of noise iterations), $j = 1$
\For{$t = 1$ to $l$}
    \State Generate decomposed features: $\boldsymbol{\mathcal{Z}}_t = \text{VMD}(\boldsymbol{\mathcal{X}} + \mathcal{N}_t(0,\hat{\sigma}^2))$
\EndFor
\For{epoch $= 1$ to epochs}
    \State Train model: $\Theta = \text{train}(\Theta, \boldsymbol{\mathcal{Z}}_j)$
    \If {$j = l$}
        \State Reset $j = 1$
    \Else
        \State Increment $j = j + 1$
    \EndIf
\EndFor
\State \textbf{Output:} Trained model parameters $\Theta$
\end{algorithmic}
\end{algorithm}

\subsection{Ablation Study}
We also assess the robustness of our model across different scenarios. In Case I, we investigate the sensitivity of various parameters within the decomposition block. Cases II and III focus on evaluating different feature parsing techniques. In Case II, modes with a mean SNR value below $-6$ dB are truncated from the feature vector. In Case III, these modes are replaced with zeros. Additionally, in Case IV~(elaborated in Algorithm~\ref{alg:1}), the decomposed modes generated by adding multiple ensembles of the noise $\mathcal{N}_t(0,\hat{\sigma}^2)$ are included in model training while keeping the same standard deviation value.  
\begin{table*}[!t]
\centering
\caption{ Performance evaluation of metrics MAE, MAPE, and RMSE on SD region for  $(\hat{\sigma}=0.1)$. $^{*}$ represents the mean of the reconstruction loss $E_{\rm R}$ given  in \eqref{eq:E_loss}.}
\resizebox{.95\textwidth}{!}{
\begin{tabular}{c|c|c|c|ccc|ccc|ccc|ccc}
\toprule
 Case &Description &Hyper-Parameters &  $E_R$ & \multicolumn{3}{c}{Horizon 3} & \multicolumn{3}{c}{Horizon 6} & \multicolumn{3}{c}{Horizon 12} & \multicolumn{3}{c}{Average}  \\
\cmidrule(r){5-16}
Scenario&& &  &MAE & RMSE & MAPE& MAE & RMSE& MAPE & MAE & RMSE & MAPE& MAE & RMSE & MAPE \\
\midrule
\multirow{4}{*}{Case I (hyper-parameters)} &
\multirow{4}{*}{Ours}
&$\alpha=2000, \epsilon=10^{-7}$& $1.773\times10^{-2}$&  11.17 & 19.18 & 10.50\% & 13.68 & 22.47 & 12.03\% & 22.00 & 77.60 & 19.01\% & 15.04 & 37.17 & 13.23\%\\
&&$\alpha=2000, \epsilon=10^{-6}$ & $1.772\times10^{-2}$ & \textbf{\textcolor{blue}{10.72}} & \textbf{\textcolor{blue}{14.87}} & \textbf{\textcolor{blue}{9.95\%}} & \textbf{\textcolor{blue}{12.86}} & \textbf{\textcolor{blue}{18.67}} & \textbf{\textcolor{blue}{11.38\%}} & \textbf{\textcolor{blue}{18.15}} & \textbf{\textcolor{blue}{27.07}} & \textbf{\textcolor{blue}{15.84\%}} & \textbf{\textcolor{blue}{13.47}} & \textbf{\textcolor{blue}{19.54}} & \textbf{\textcolor{blue}{12.06\%}} \\
&&$\alpha=1000, \epsilon=10^{-7}$ & $1.818\times10^{-2}$  & 11.10 & 17.09 & 10.70\% & 14.43 & 30.82 & 14.23\% & 23.19 & 75.62 & 22.48\% & 15.73 & 41.13 & 15.28\% \\
&&$\alpha=1000, \epsilon=10^{-6}$ & $1.817\times10^{-2}$ & 10.73 & 15.20 & 10.47\% & 13.45 & 25.75 & 13.24\% & 21.45 & 78.37 & 22.85\% & 14.68 & 38.95 & 14.96\% \\
\midrule
\multirow{2}{*}{Case-II (truncation 8 modes)}
&VMGCN &$\alpha=2000, \epsilon=10^{-7}$ & $2.12 \times 10^{-2}$ & 13.33 & 20.46 & 11.36\% & 15.31 & 27.34 & 13.14\% & 22.28 & 87.22 &21.40\% & 16.37 & 41.74 & 14.63\% \\
&Ours&$\alpha=2000, \epsilon=10^{-7}$ & $2.12 \times 10^{-2}$ & \textbf{\textcolor{blue}{13.09}} & \textbf{\textcolor{blue}{19.51}} & \textbf{\textcolor{blue}{10.96\%}} & \textbf{\textcolor{blue}{14.40}} & \textbf{\textcolor{blue}{21.62}} & \textbf{\textcolor{blue}{11.82\%}} & \textbf{\textcolor{blue}{18.17}} & \textbf{\textcolor{blue}{28.69}} & \textbf{\textcolor{blue}{15.15\%}} & \textbf{\textcolor{blue}{14.75}} & \textbf{\textcolor{blue}{22.48}} & \textbf{\textcolor{blue}{12.16\%}} \\
\midrule
\multirow{2}{*}{Case-III (zeros)}
&VMGCN&$\alpha=2000, \epsilon=10^{-7}$ & $1.79 \times 10^{-2}$ & \textbf{\textcolor{blue}{10.62}} & \textbf{\textcolor{blue}{14.58}} & \textbf{\textcolor{blue}{10.59\%}} & \textbf{\textcolor{blue}{12.45}} & \textbf{\textcolor{blue}{17.85}} & \textbf{\textcolor{blue}{11.81\%}} & \textbf{\textcolor{blue}{17.61}} & \textbf{\textcolor{blue}{25.88}} & \textbf{\textcolor{blue}{15.97\%}} & \textbf{\textcolor{blue}{13.10}} & \textbf{\textcolor{blue}{18.81}} & \textbf{\textcolor{blue}{12.32\%}} \\
&Ours&$\alpha=2000, \epsilon=10^{-7}$ & $1.79 \times 10^{-2}$ & 11.19 & 16.66 & 10.73\% & 14.14 & 27.88 & 13.84\% & 22.83 & 83.68 & 24.26\% & 15.44 & 41.20 & 15.61\% \\
\midrule
\multirow{2}{*}{Case-IV (time varying noise)}
&VMGCN &$\alpha=2000, \epsilon=10^{-7}$ & $1.78 \times 10^{-2}$$^{*}$ & 11.26 & 17.69 & 12.09\% & \textbf{\textcolor{blue}{13.39}} & 27.65 & \textbf{\textcolor{blue}{13.40\%}} & \textbf{\textcolor{blue}{19.86}} & \textbf{\textcolor{blue}{67.03}} & \textbf{\textcolor{blue}{19.71\%}} & \textbf{\textcolor{blue}{14.29}} & \textbf{\textcolor{blue}{34.99}} & 14.57\% \\
&Ours&$\alpha=2000, \epsilon=10^{-7}$ & $1.78 \times 10^{-2}$$^{*}$ & \textbf{\textcolor{blue}{10.94}} & \textbf{\textcolor{blue}{15.98}} & \textbf{\textcolor{blue}{10.51\%}} & 13.94 & \textbf{\textcolor{blue}{26.64}} & 13.53\% & 21.46 & 72.95 & 20.60\% & 14.97 & 37.15 & \textbf{\textcolor{blue}{14.34\%}} \\

\hline
\end{tabular}
}
\label{tab:hyperparameter}
\end{table*}

Table~\ref{tab:hyperparameter} Case I describes the effectiveness of bandwidth constraint $\alpha$ and tolerance convergence $\epsilon$ on the performance metrics for the prediction metrics on the SD region. In this comparison, the data is contaminated with the noise $\hat{\sigma}=0.1$. For evaluating the performance, we define the reconstruction error~(loss) ($E_{\rm R}$) defined as
\begin{equation}
\label{eq:E_loss}
    E_{\rm R}=\frac{1}{TN} \sum\limits_{n=1}^{N}  \sum\limits_{t=1}^{T}(y^n(t)-\sum\limits_{i=1}^{K} U_i^n(t))^2,
\end{equation}
\noindent where $y^n(t)$ and $U_i^n(t)$ represent the noisy signal and mode in the time domain for $n^{\rm{th}}$ node, respectively. 
The reconstruction term is relatively low for a large bandwidth value $\alpha$, indicating that the residual value decreases as the bandwidth constraint increases. Generally, higher convergence $\epsilon$ values lead to better prediction performance.~\cite{Ahmad2024variational} shows that the lower value of bandwidth constraint encourages better performance for the noise-free case. However, in the presence of additive noise, a higher bandwidth constraint yields more accurate predictions.

The performance of our model is superior with respect to all metrics to VMGCN when the 8 high-frequency modes are truncated and the features primarily consist of low-frequency modes in Case II are used. It indicates that channel attention is considered to highlight the important features rather than to suppress the noise. The overall performance of our model degrades Case III when the noisy features are replaced by zero values, it depicts that our model is sensitive to feature representation compared to VMGCN. With the multiple-ensemble noise, VMGCN overall performs better as compared to our model. Our model enables better short-term predictions but the performance degrades on the long-term horizon predictions.  

\begin{figure}[!t]
\centering
\begin{subfigure}[b]{0.15\textwidth}
     \centering
    \includegraphics[width=\linewidth]{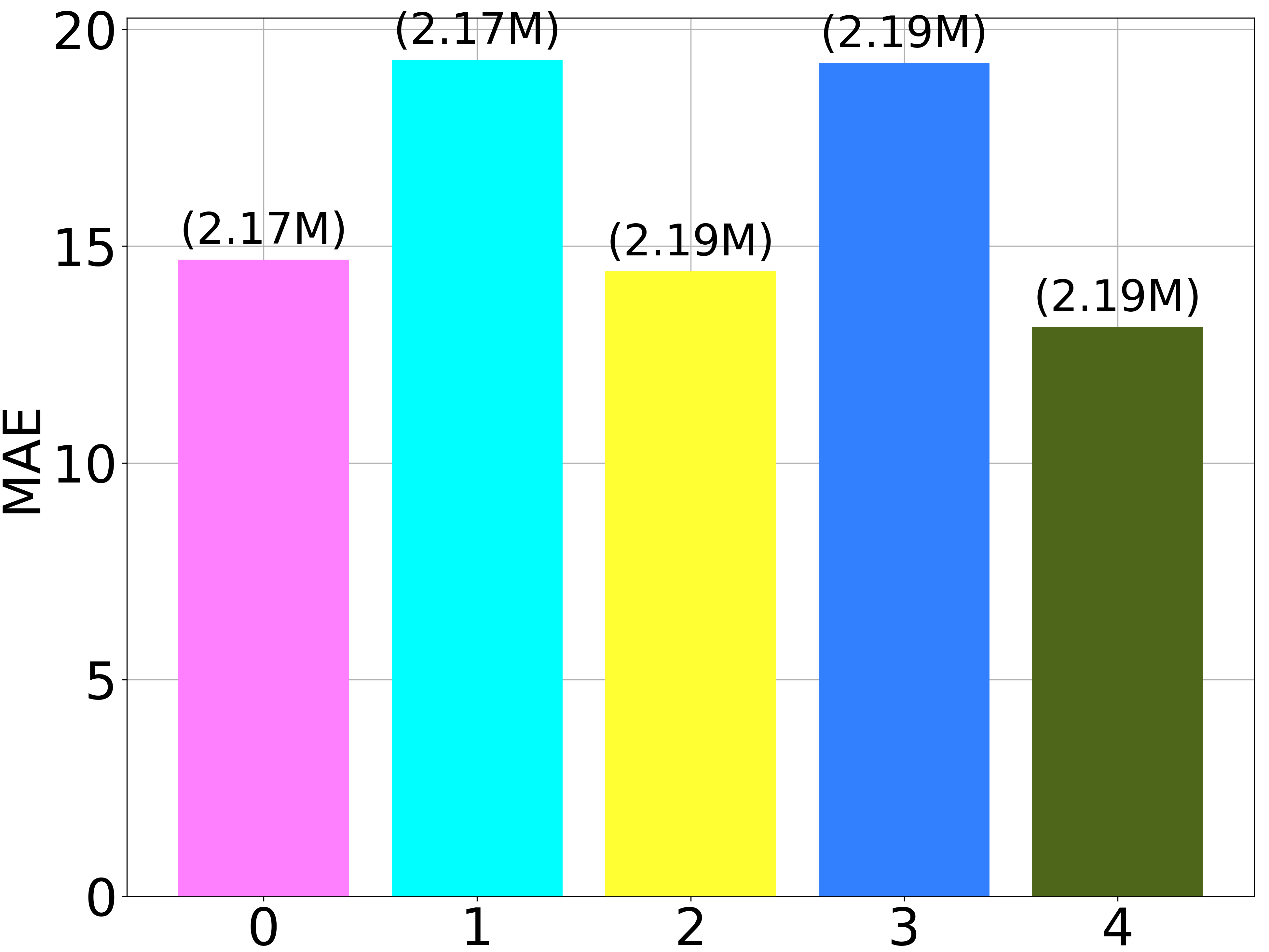}
    \caption{}
\end{subfigure}
\hfill
\begin{subfigure}[b]{0.15\textwidth}
     \centering
    \includegraphics[width=\linewidth]{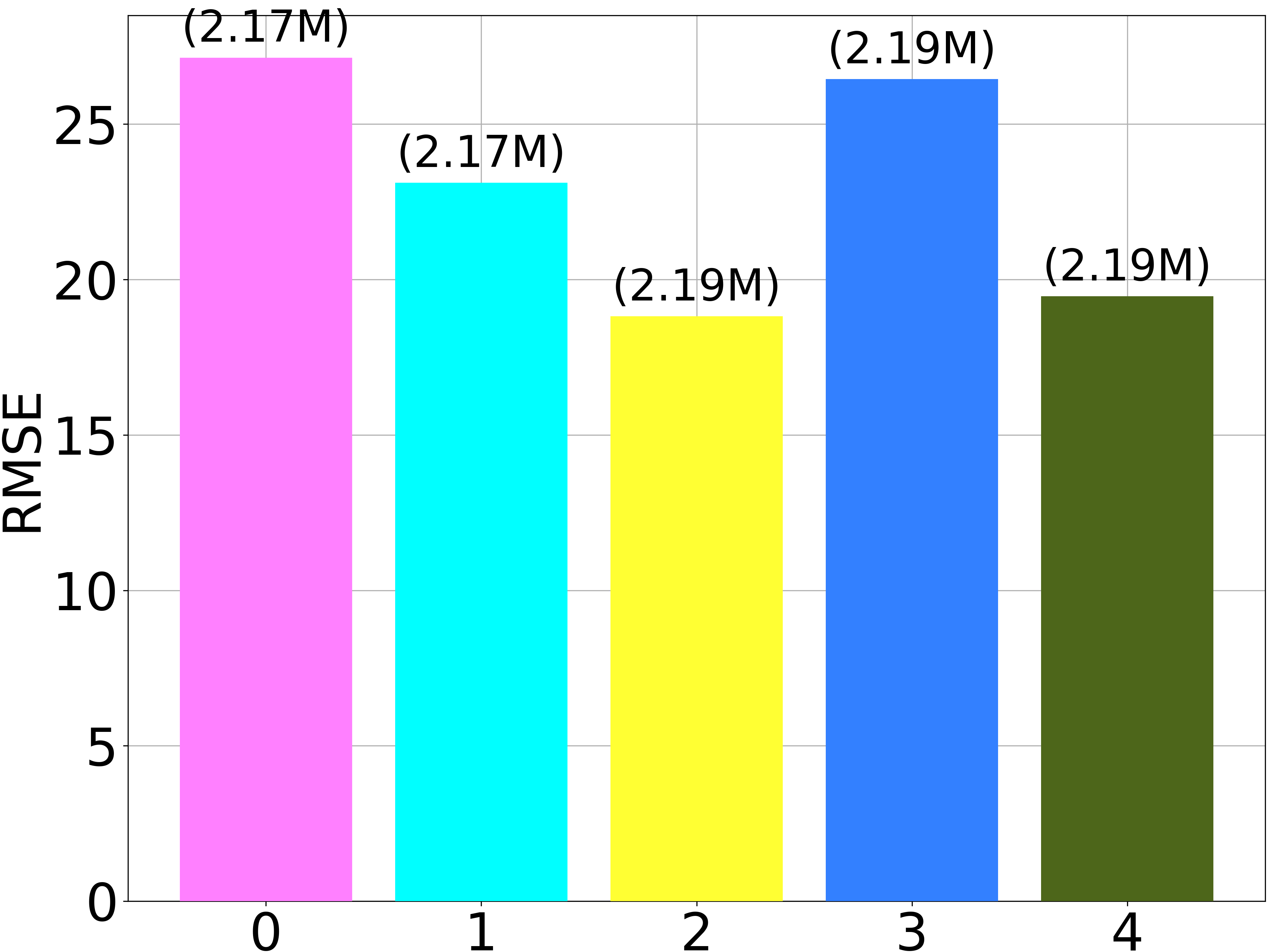}
    \caption{}
\end{subfigure}
\hfill
\begin{subfigure}[b]{0.15\textwidth}
     \centering
    \includegraphics[width=\linewidth]{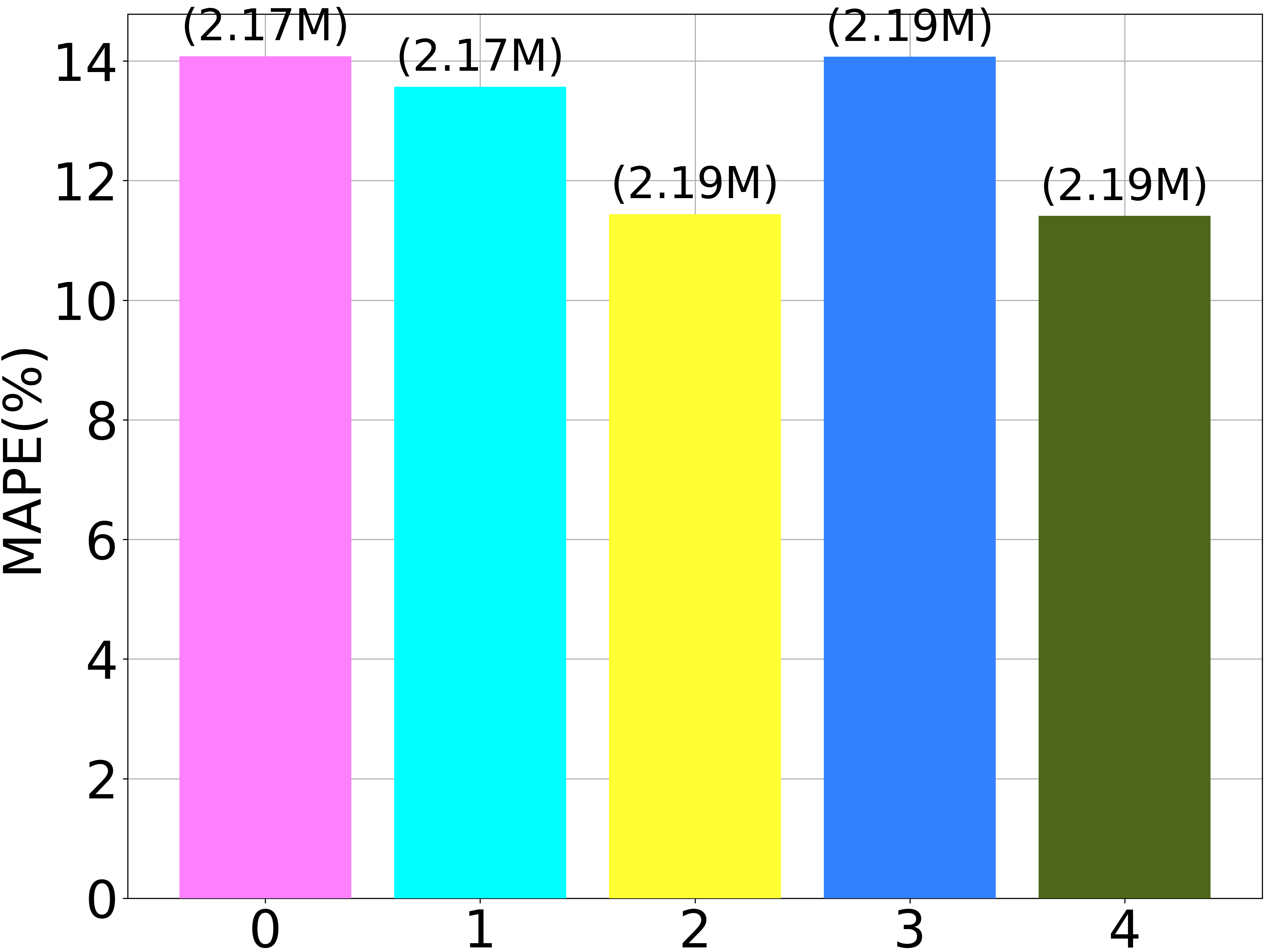}
    \caption{}
\end{subfigure}
\vspace{-1mm}
   \caption{Performance metrics (a) MAE, (b) RMSE, and (c) MAPE on the average horizon for different combinations of hyperparameters. Different combinations of hyperparameters in the sequence (B, F, M) are indicated, where B, F, and M denote the batch size, the number of Chebyshev filters, and the order of polynomial, respectively. 0: (48,32,2), 1: (48, 32, 3), 2: (32, 64, 3), 3: (48, 64, 2), and 4: (48, 64, 3).
   }
   \label{fig:hyperparameters}
\end{figure}

We also analyze the performance of the proposed model for different combinations of hyperparameters such as the size for mini-batch (B), Chebyshev filters (F), and the order of polynomials (M) in the training pipeline for the SD region on $\hat{\sigma}=0.1$, replacing the highest frequency mode with zero value. The error metrics for different combinations are plotted in Fig.~\ref{fig:hyperparameters}, where we observe that the hyperparameters (48, 64, 3) correspond to the (B, F, M) providing the better performance, primarily due to an increase in the number of parameters. 
\section{Conclusion}
\label{sec:conclusion}
In this work, we have introduced a novel approach for enhancing the robustness and accuracy of spatiotemporal long-term predictions in the presence of noise by leveraging a variational mode graph neural network (VMGCN) with a 3D channel attention mechanism. Our framework decomposes corrupted signals into modes and integrates spatial, temporal, and channel attention to isolate and emphasize the meaningful patterns within the data while suppressing noise. The inclusion of a learnable soft thresholding technique and a signal-to-noise ratio (SNR)-based feature reduction method further refines the model's ability to differentiate between important and irrelevant modes. Our extensive experiments on the LargeST dataset, combined with various noise levels, demonstrate that the proposed architecture outperforms baseline models in terms of long-term prediction accuracy.  Furthermore, our analysis reveals that the proposed architecture is resilient to noise, sensitive to feature representation, and effective mode truncation. Future work may explore alternative noise models and additional datasets to evaluate the generalization capability of the proposed approach. These findings offer valuable insights for future research on improving the reliability and performance of prediction models in complex, noisy environments, paving the way for further advancements in time-series prediction across a wide range of applications including traffic forecasting.
\section{Acknowledgment}
\vspace{-0.2em}
 This work is supported by the Higher Education Commission of Pakistan under its National Research Grant for Universities (NRPU) scheme (Project No. 16555).


\scriptsize
\bibliographystyle{IEEEbib}
\bibliography{aaai25,refs}

\end{document}